\documentclass[sigconf]{acmart}
\usepackage{color}

\usepackage{booktabs} % For formal tables
\usepackage{amsmath}
\usepackage{bm}
\usepackage{graphicx}
\usepackage{subcaption}
\usepackage{physics}
\usepackage{booktabs}
\usepackage{multirow}
\usepackage{tablefootnote}

\usepackage{mathtools}
\usepackage{standalone}
\usepackage{booktabs, dcolumn}
\usepackage{pdflscape}
\usepackage{booktabs, multirow}
\usepackage[flushleft]{threeparttable}
\setcopyright{rightsretained}
\acmConference[XXX]{XXX}{XXX}
\acmArticle{4}
\acmPrice{15.00}

\usepackage{xcolor}

\begin{document}
	\title[Improving Temporal Interpolation of Head and Body Pose]{Improving Temporal Interpolation of Head and Body Pose using Gaussian Process Regression in a Matrix Completion Setting}
	%\titlenote{Produces the permission block, and
	%  copyright information}
	%\subtitle{Extended Abstract}
	%\subtitlenote{The full version of the author's guide is available as
	%  \texttt{acmart.pdf} document}

	\author{Stephanie Tan}
	%\authornote{Dr.~Trovato insisted his name be first.}
%	\orcid{1234-5678-9012}
	\affiliation{
	  \institution{TU Delft}
	%  \streetaddress{P.O. Box 1212}
%	  \postcode{43017-6221}
	}
	\email{S.Tan-1@tudelft.nl}
	
	\author{Hayley Hung}
	%\authornote{The secretary disavows any knowledge of this author's actions.}
	\affiliation{%
	  \institution{TU Delft}
	%  \streetaddress{P.O. Box 1212}
	}
	\email{H.Hung@tudelft.nl}
	
%	\author{Anonymous Authors}
	%\authornote{The secretary disavows any knowledge of this author's actions.}
	%\affiliation{%
	%	\institution{X}
	%  \streetaddress{P.O. Box 1212}
	%\email{H.Hung@tudelft.nl}

	\begin{abstract}
		This paper presents a model for head and body pose estimation (HBPE) when labelled samples are highly sparse. 
		% Such sparse label settings are highly relevant when working with multimodal data (such as video and wearable sensors) which are likely to be incomplete. 
		The current state-of-the-art multimodal approach to HBPE  utilizes the matrix completion method in a transductive setting to predict pose labels for unobserved samples. 
		Based on this approach, the proposed method tackles HBPE when manually annotated ground truth labels are temporally sparse. We posit that the current state of the art approach oversimplifies the temporal sparsity assumption by using Laplacian smoothing. Our final solution uses : i) Gaussian process regression in place of Laplacian smoothing, ii)  head and body coupling, and iii) nuclear norm minimization in the matrix completion setting. The model is applied to the challenging SALSA dataset for benchmark against the state-of-the-art method. Our presented formulation outperforms the state-of-the-art significantly in this particular setting, e.g. at 5\% ground truth labels as training data, head pose accuracy and body pose accuracy is approximately 62\% and 70\%, respectively. As well as fitting a more flexible model to missing labels in time, we posit that our approach also loosens the head and body coupling constraint, allowing for a more expressive model of the head and body pose typically seen during conversational interaction in groups. 
		This provides a new baseline to improve upon for future integration of multimodal sensor data for the purpose of HBPE. 
		%\begin{itemize}
		%	\item \TODO{Talk} about lack of annotated multimodal data, and in this case, difficulty in obtaining labels due to the nature of video data and social scene (crowdedness--> self-occlusion, lighting, camera view), \item need to work with limited number of labels. 
		%	\item This work focuses on estimating head and body pose in crowded social scene scenario using gaussian process regression that is able to recover a relatively large percentage of missing labels in large continuous time segment. 
		%	\item Reported results on the SALSA dataset and have shown desired result
		%	\item Approx 0.5 column 
		%\end{itemize}
		
	\end{abstract}
	
	%\begin{CCSXML}
	%<ccs2012>
	%  <concept_id>10010520.10010553.10010562</concept_id>
	%  <concept_desc>Computer systems organization~Embedded systems</concept_desc>
	%  <concept_significance>500</concept_significance>
	%  <concept_id>10010520.10010575.10010755</concept_id>
	%  <concept_desc>Computer systems organization~Redundancy</concept_desc>
	%  <concept_significance>300</concept_significance>
	%  <concept_id>10010520.10010553.10010554</concept_id>
	%  <concept_desc>Computer systems organization~Robotics</concept_desc>
	%  <concept_significance>100</concept_significance>
	%  <concept_id>10003033.10003083.10003095</concept_id>
	%  <concept_desc>Networks~Network reliability</concept_desc>
	%  <concept_significance>100</concept_significance>
	%</ccs2012>
	%\end{CCSXML}
	
	%\ccsdesc[500]{Computer systems organization~Embedded systems}
	%	\ccsdesc {are these mandatory?}

	%\keywords{ACM proceedings, \LaTeX, text tagging}

	\maketitle
	
	\section{Background}
\begin{figure}
		\centering
		\begin{subfigure}[b]{0.24\columnwidth}
			\includegraphics[width=\columnwidth]{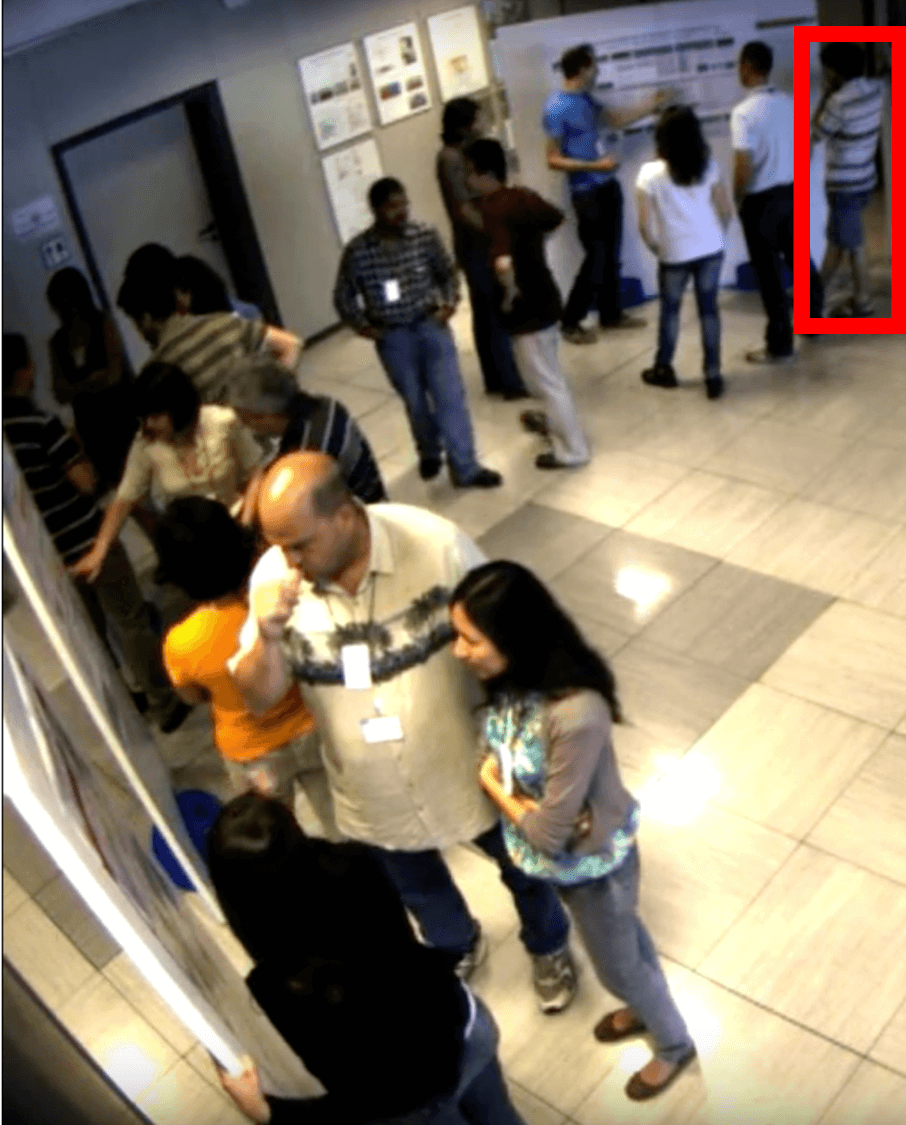}
			\caption{}
			\label{fig:lowresolution}
		\end{subfigure}
		~ %add desired spacing between images, e. g. ~, \quad, \qquad, \hfill etc. 
		\begin{subfigure}[b]{0.24\columnwidth}
			\includegraphics[width=\columnwidth]{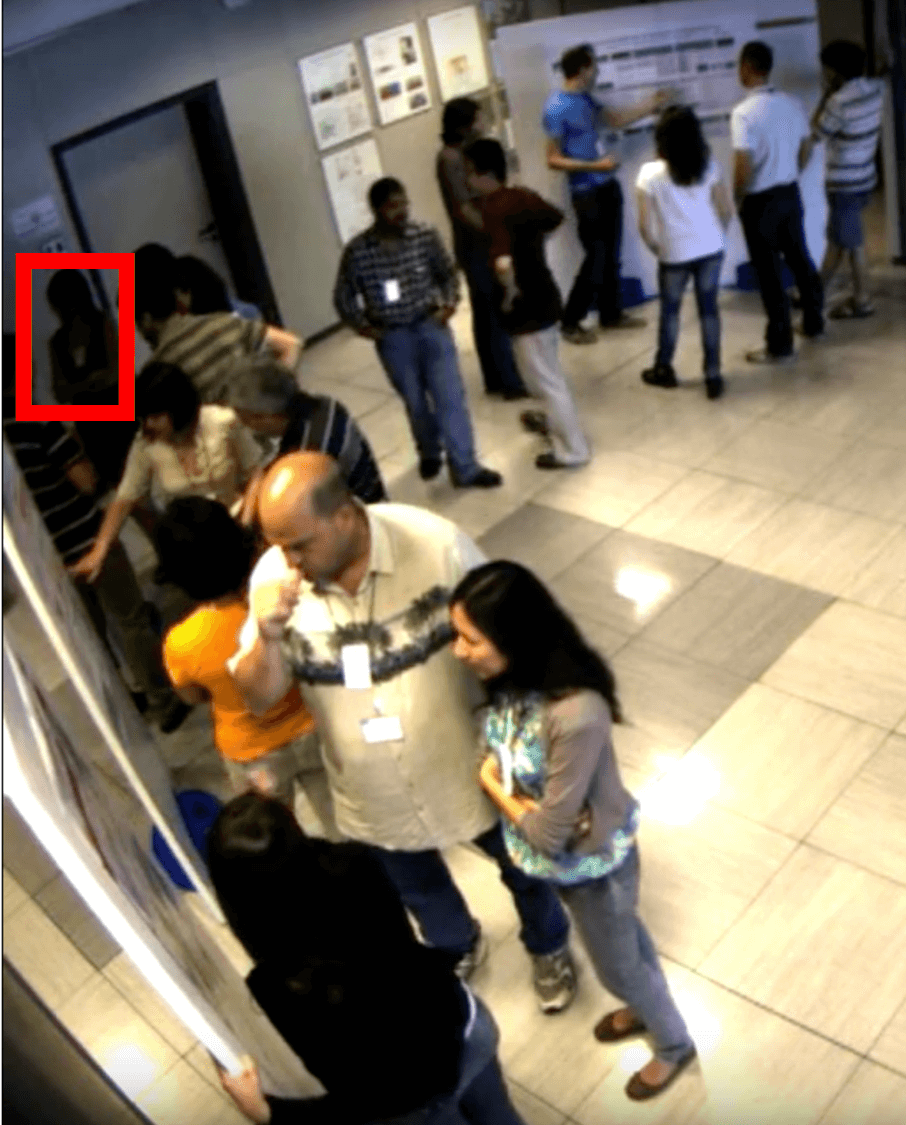}
			\caption{}
			\label{fig:lowvisibilityr}
		\end{subfigure}
		~ %add desired spacing between images, e. g. ~, \quad, \qquad, \hfill etc. 
		\begin{subfigure}[b]{0.24\columnwidth}
			\includegraphics[width=\columnwidth]{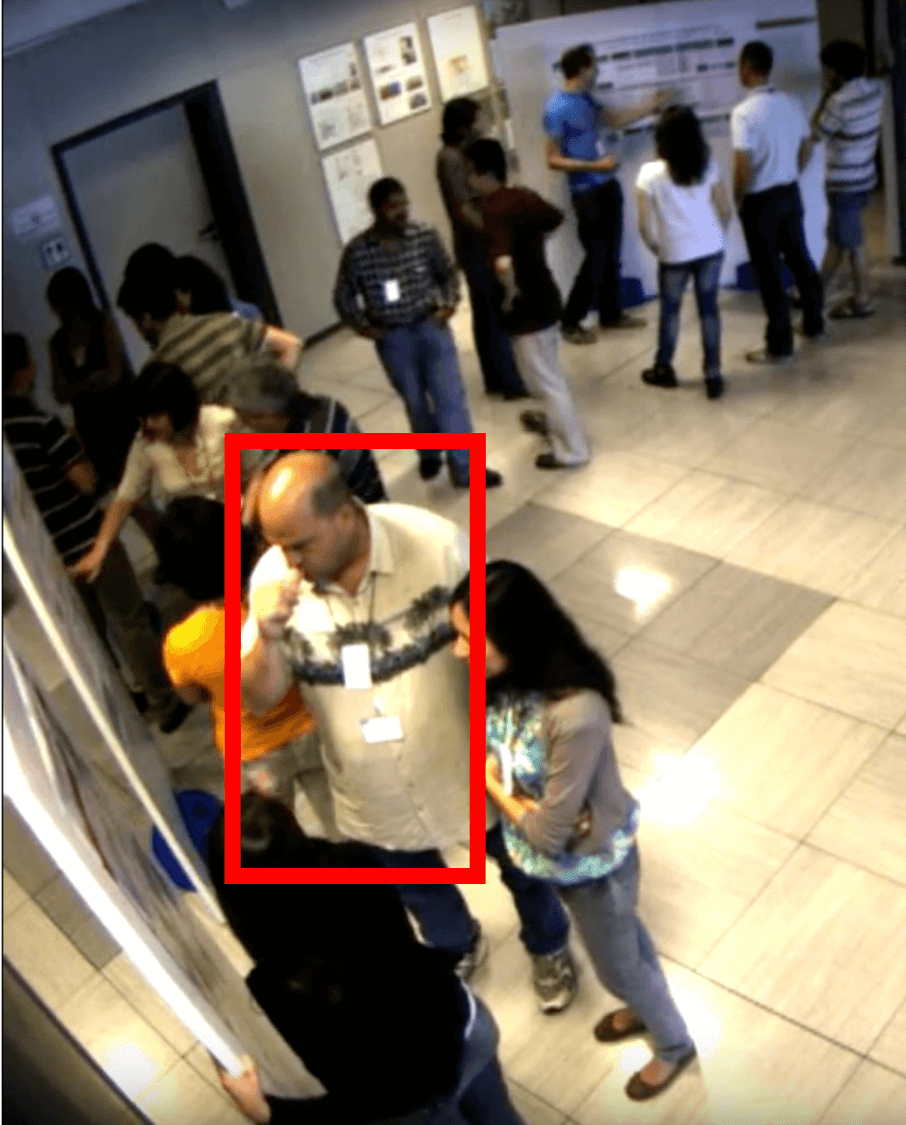}
			\caption{}
			\label{fig:clutter}
		\end{subfigure}
		\begin{subfigure}[b]{0.24\columnwidth}
			\includegraphics[width=\columnwidth]{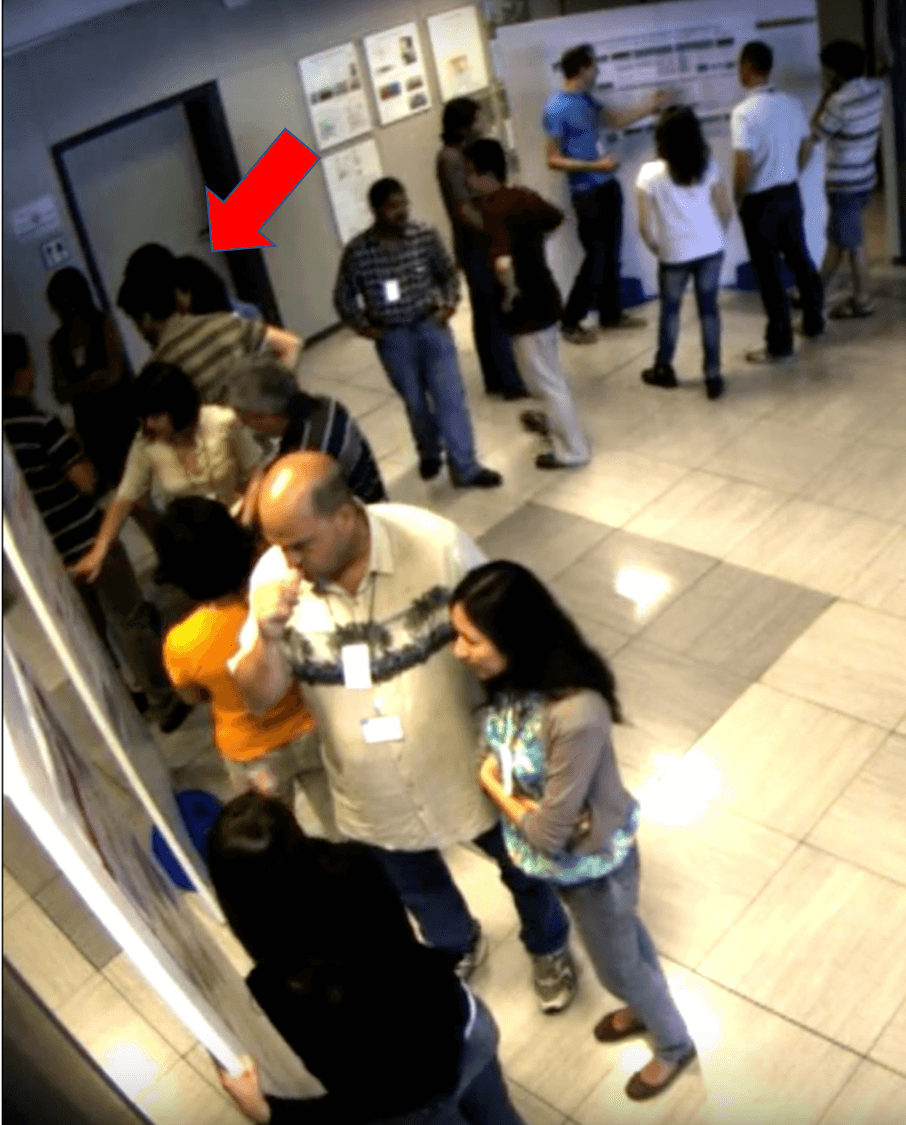}
			\caption{}
			\label{fig:occlusion}
		\end{subfigure}
		
		\caption{Examples of HBPE challenges from the SALSA dataset \cite{AlamedaPinedaEtAl2016}. (a) Low resolution (b) Low visibility (c) background clutter (d) occlusion}\label{scene}
	\end{figure}

	Pose estimation has been a popular subject of interest within the computer vision community. While deep learning based state-of-the-art pose estimation methods  \cite{GulerEtAl2018, WeiEtAl2016, ToshevEtAl2014, TompsonEtAl2014} have achieved remarkable results in articulated pose estimation (i.e. detection and prediction of the location of body parts and joints), pose estimation remains challenging particularly for crowded scenes in the surveillance setting. Hence, it is limited to only head and body pose estimation (HBPE). Despite the seeming simplification of the task, challenges of HBPE in this particular setting \cite{HuEtAl2004} include but are not limited to low resolution, low light visibility, background clutter and occlusions (see Figure \ref{scene} for example).
	
	HBPE is traditionally a vision-only task. But to tackle these challenges, researchers can leverage on a multi-view camera and multi-sensor scenario \cite{AlamedaPinedaEtAl2016}. The multi-view camera setting provides multiple perspectives of people in the scene to acquire better HBPE. More interestingly, wearable sensors such as microphones, infrared or bluetooth proximity sensors, etc. have shown an ability to recover HBPE independent from the video modality \cite{KokEtAl2017}. Additionally, they can provide more  fine-grained information of the human subjects that would not otherwise be available from video only. More specifically, studying small group interactions in crowded space can benefit from data of multiple modalities \cite{Gatica-PerezEtAl2009}. In combination with video, these wearable sensors provide a multimodal platform to study detailed and rich information about the human subjects by complementing and enhancing HBPE, which is particularly crucial to the analysis of group and crowd behavior. 
	%\HH{I think you need to add something just before this stentence, or perhaps earlier to motivate why you specically want to study this setting. Otherwise it could be strange to be only testing your method on the SALSA data and not others....}
	
	Even though it would be ideal to combine multiple modalities, wearable sensors such as microphones and infrared proximity sensors which have previously been used to study group interaction and behavior, are significantly less reliable and noisier compared to surveillance video footage for the purpose of HBPE. Another problem is that malfunctions of wearable sensors are more difficult to notice compared to those of video cameras, especially during real-time data collection where there may be visual confirmation of camera functionality but not of wearable sensor data. Due to the difficulties of working with wearable sensors, the resulting data can be either partial or entirely missing \cite{HiggerEtAl2013}. Given that working with a patchwork of multimodal data can be hard, in this paper, we exploit them as part of an initialization step and focus on the problem of interpolating between sparse labels. 
	%When such scenario occurs, the question boils down to exploring the possibility of HBPE only using visual information, when sensor modality fails to provide even noisy estimates of head and body poses due to missing data. 
	
	% \HH{These remaining two sentences need to be removed... we put ourselves in very dangerous territory by talking about a vision only solution.}
	% Though it has already been shown that HBPE in the group interaction setting can be improved by leveraging information from multiple modalities, this study simulates the realistic situation when other modalities fail to record data or provide even noisy estimates of HBPE. 
	% Rather than recovering the missing data from the sensor modalities, this study targets the possibility of HBPE by only using visual information for video cameras, as there is still room for improvement by considering HBPE as a computer vision task in this setting. 
	
	% This in practices leads to faster convergence of the matrix completion problem and leads to a slight improvement in performance.

	The setting of this study is that: i) there is a relatively small number of head and body pose samples ($\sim 10^2-10^3$) for each subject, ii) we want to predict pose labels for unobserved samples only using a very small number ($\sim5\%$) of sparsely distributed ground truth labels, and iii) we want to take advantage of the temporal structure within the pose label data. A deep learning based method that takes into account this setting will perform sub-optimally due to small number of training samples, and also require extensive computational power and hyperparameter tuning. On the other hand, a matrix completion based transductive learning method which is more explainable and less computationally expensive, addresses the problem setting adequately. Inspired and building upon previous work by  \citet{AlamedaPinedaEtAl2015}, the contributions of this study are: i) an enhanced temporal smoothing scheme based on Gaussian process regression for label propagation, and ii) a more interpretable person-wise pose label prediction implementation in the transductive setting using matrix completion.
	
	\section{Related work}
	Head pose estimation (HPE) and body pose estimation (BPE) have been primarily studied by the computer vision community \cite{SigalEtAl2014}. While impressive results could be achieved using end-to-end deep learning architectures when data capturing frontal faces \cite{MurphyEtAl2009} or  the full body \cite{CaoEtAl2017}, HPE and BPE remain to be challenging tasks when dealing with wide angle surveillance, with low resolution, heavy occlusions of targets, and cluttered backgrounds. The problem is often reduced to an 8-class classification problem (dividing $360^{\circ}$ into eight sectors), though formulating HBPE as a regression problem \cite{VaradarajanEtAl2018} or being able to reduce coarseness in estimations can provide more meaningful information for higher level social tasks, such as predictions of social attention direction \cite{MasseEtAl2017} and personality traits \cite{SubramanianEtAl2013}.
	Pioneering work \cite{SigalEtAl2006, BaEtAl2009, ChenEtAl2011}  in HPE and BPE saw first successes of these tasks based on probabilistic frameworks (e.g. dynamic Bayesian networks, hidden Markov models, etc.). Due to the physical constraint of relative head and body pose and a person's direction of movement, one line of work focuses on the joint estimation of head and body pose to achieve improved results \cite{ChenEtAl2011}. Overall, there are more previous works on HPE compared to BPE in the surveillance and crowded space setting. In this particular setting, human heads can be more easily seen and HPE typically already contains rich enough information for high level tasks \cite{BaEtAl2009}. On the other hand, humans bodies are usually occluded because of the camera angle from the top, which makes it more difficult to predict body orientations without side information such as walking direction, etc. In contrast, HBPE in other contexts such as AR/VR video gaming, sports, etc. where full body poses data are captured by frontal view camera, is much more well-studied and can be represented by a considerable number of work (e.g. \cite{CaoEtAl2017},\cite{InsafutdinovEtAl2016}). 
	% This is because researchers have channeled their efforts towards estimating full body poses, by estimating \ST{joint locations} \HH{you mean estimating joint locations?}, rather than estimating body pose based on shoulder or hip orientation \HH{cite? I'm also not sure the 'because' part of this sentence really explains why people have focused on these scenarios. Perhaps it's better to rephrase by talking first about the settings that are typically popular in these domains and therefore where the camera would be relative to the person? }
	%\HH{I'm not sure I get what you are trying to say in this last bit of the sentence. How it a crowded social scene a choice... I think perhaps you are trying to say too many things in one sentence?} , which in crowded social scenes is a choice limited by the scenario and type of data. 
	Additionally, the line of works on low-resolution HPE leverages on multi-view surveillance images. \citet{HasanEtAl2018} have recently proposed a noteworthy deep learning method based on Long Short Term Memory (LSTM) neural networks to jointly forecast trajectories and head poses. This work points to the possibility of utilizing LSTM models in predicting head and body pose sequences, which is more informative compared to solving HPE and BPE in a classification setting using Convolutional Neural Networks (CNN) \cite{LuEtAl2016}.
	
	% \subsection{HBPE using matrix completion}
	In this paper, we propose to use matrix completion for HBPE, which was first proposed by \citet{AlamedaPinedaEtAl2015}. This approach combines head and body visual features, inferred head orientation labels from audio recordings, body orientation labels from infrared proximity sensors, and 
	%\HH{partial ?}
	manually annotated labels of some but not all frames. To reduce the manual effort of annotating the head pose, labels were only created every 3 seconds. Alameda-Pineda et al. poses the estimation of head and body orientations as a matrix completion problem where the visual features and labels from either wearable sensors or manual annotations are concatenated into a heterogeneous matrix, for head and body respectively. Due to sparsity and noise in the data extracted from the wearable sensors, the underlying challenge is to construct a matrix that is temporally smooth; and that is consistent with the manual annotations, the observed wearable
	%\HH{you mean both the visual and sociometer readings? traditionally sensor readings could be read as being related to the wearables and not the visual modality}
	sensor readings, and the physical constraints that tend to couple the head and body behaviour together. 
	
	\section{Our Approach}
\begin{figure}
		\includegraphics[width=\columnwidth]{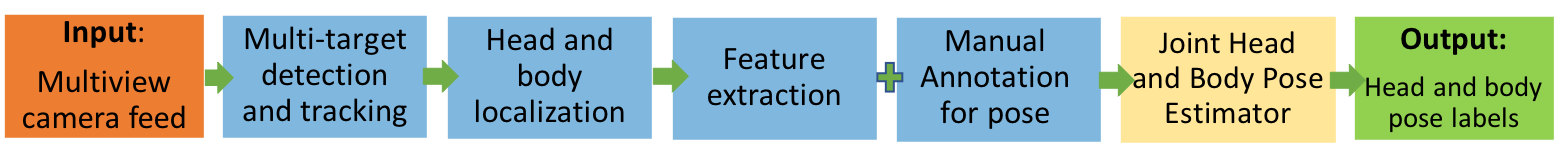}
		\caption{Overall work flow of this study. The focus of this study is highlighted in yellow}\label{flow}
	\end{figure}

\begin{figure}
		\includegraphics[width=\columnwidth]{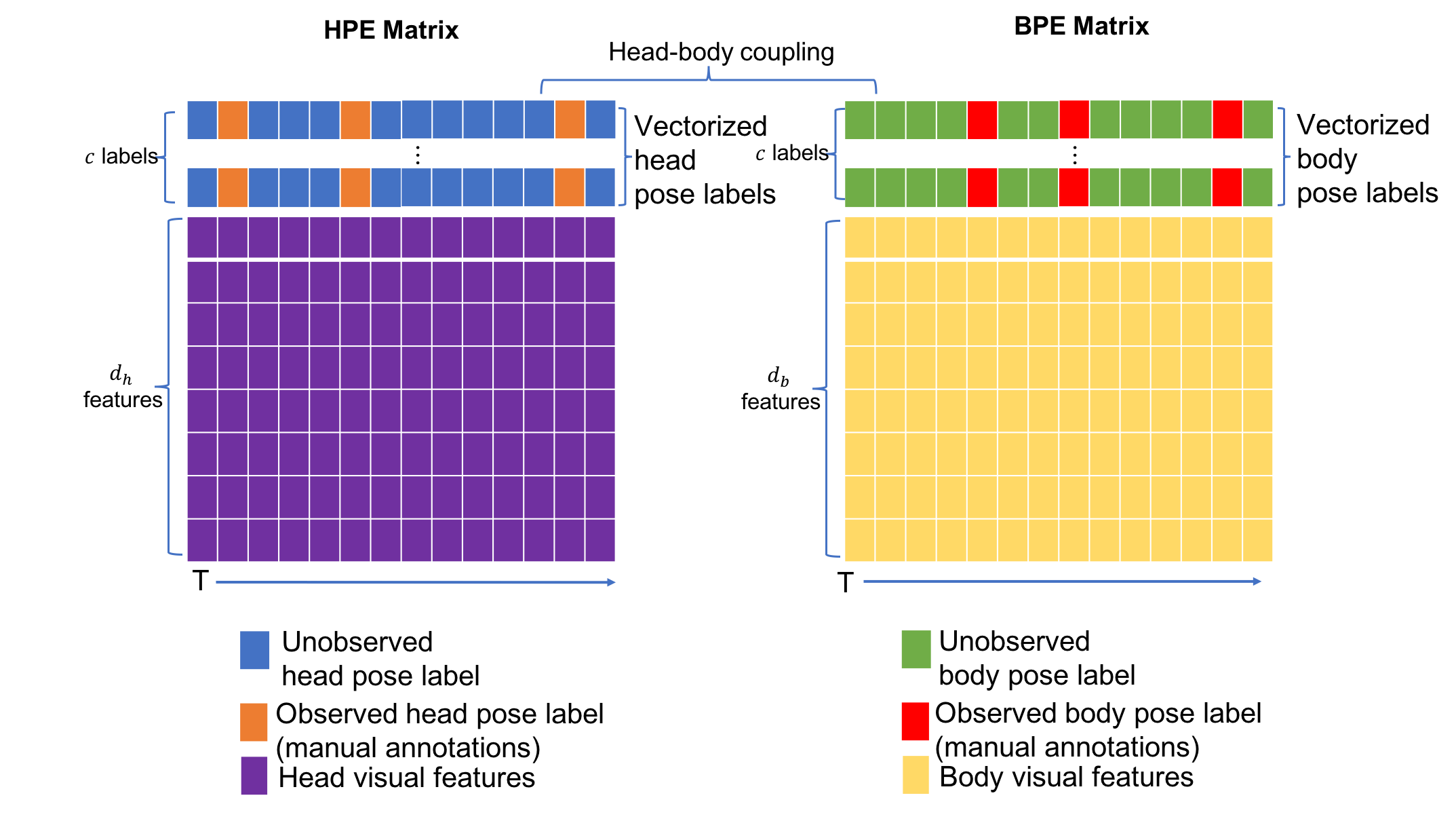}
		\caption{Graphical representation of the features and labels heterogeneous matrix}\label{mc}
	\end{figure}

	The scope of the study is to jointly predict head and body pose labels as an 8-class classification problem (dividing $360^{\circ}$ into eight sectors) in a matrix completion transductive learning setting. Before performing HBPE, upstream processes such as multi-person detection and tracking in videos, head and body localization, and appearance-based visual feature extraction are carried out as outlined in Figure \ref{flow} \cite{ChenEtAl2011, AlamedaPinedaEtAl2015}. 
	The construction of a matrix consisting of visual features and manually annotated labels is illustrated in Figure \ref{mc}. Head pose features and labels are arranged into one such matrix, and similarly for body pose features and labels. Head and pose labels of each participant (independent of other participants) are estimated by completing their head and body matrices jointly. The technical core of constructing such matrices for HBPE and jointly completing the head and body matrices using our formulation is discussed in Section 4, followed by details on experimental conditions pertaining to the upstream processes (see blue modules in Figure \ref{flow}) in Section 5.
	
	%\begin{itemize}
	%	\item small section in words explaining the upstream processes (detection, visual extraction, etc, mention details can be found in Xavi's paper)
	%	\item flow chart indicating the workflow, not too much detail, basically just clarify inputs to the matrix completion. 
	%	\item approx 0.5-0.75 column
	%	\item Approach section usually also gives a map of how the remaining sections map to the different modules of the approach.
	%\end{itemize}

	\section{Methodology}
	In the supervised learning setting for a linear classifier,  the objective is to learn the weight matrix  $\bm{W}\in\mathbb{R}^{c\times (d+1)}$, which maps the $d$-dimensional features space $\bm{X}\in\mathbb{R}^{d\times T}$ to the $c$-dimensional (number of classes) output space $\bm{Y}\in\mathbb{R}^{c\times T}$ where $T$ denotes the number of samples in time, by minimizing the loss on a training set  $N_{\text{train}}$ as
	\begin{equation}
	\displaystyle \arg \min_{ \bm{W}} \sum_{i\in N_{\text{train}}} \text{Loss } \left( \bm{Y}_i, \bm{W} \begin{bmatrix} \bm{X}_i \\  1 \end{bmatrix} \right).
	\end{equation}
	When dealing with noisy features and fuzzy labels, previous research by \citet{GoldbergEtAl2010, CabralEtAl2011, BommaEtAl2015} have empirically shown the practicality of casting a classification problem into a transductive learning setting such as matrix completion. To that purpose, borrowing from the linear classifier setting, a heterogeneous matrix can be built by concatenating the pose labels $\bm{Y}\in \mathbb{R}^{{c}\times T}$, visual features $\bm{X} \in \mathbb{R}^{{d}\times T}$, and a row of $1$'s (to model for bias) as
	\begin{equation}
	\label{hetero matrix def}
	\bm{J} = \begin{bmatrix} \bm{Y}\\\bm{X}\\\bm{1} \end{bmatrix},
	\end{equation}
	where $\bm{J} \in \mathbb{R}^ {(c+d+1) \times T}$. Note that $\bm{Y}$ is a vectorized one hot representation of pose labels. 
	
	In the HBPE setting, the duration that we are interested in predicting the pose estimations for is indicated by $T$ and this is represented by arranging samples column-wise for temporal consistency. The number of pose classes possible is denoted by $c$. Dividing $360^{\circ}$ into eight sectors means that there are eight possible classes and each pose belongs to one of the eight classes. For example, a pose angle between $45^{\circ}$ and $90^{\circ}$ would be indicated by the vector $[0,1,0,0,0,0,0,0] ^\top  \in \mathbb{R}^{c \times 1}$. The head and body label matrices are denoted by $\bm{Y}_h\in \mathbb{R}^{c\times T}$ and $\bm{Y}_b\in\mathbb{R}^{c\times T}$ respectively. The feature matrices $\bm{X}_h\in\mathbb{R}^{d_h\times T}$ and $\bm{X}_b\in\mathbb{R}^{d_b\times T}$ contain the visual features from head and body crops of each person, where $d_h$ and $d_b$ denote the respective feature dimensionality. Following the definition in \eqref{hetero matrix def}, the heterogeneous matrices are $\bm{J}_h =  \begin{bmatrix} \bm{Y}_h^\top, \bm{X}_h^\top, \bm{1}^\top \end{bmatrix} ^\top$ and $\bm{J}_b =  \begin{bmatrix} \bm{Y}_b^\top, \bm{X}_b^\top, \bm{1}^\top \end{bmatrix} ^\top$ for head pose and body pose estimation respectively. In addition, a projection matrix $\bm{P}_h = [\bm{I}^{c \times c}, \bm{0}^{c \times (d_h+1)}]$ is introduced to extract only the head pose labels from the heterogeneous matrix $\bm{J}_h$. In a similar manner, a projection matrix $\bm{P}_b = [\bm{I}^{c \times c}, \bm{0}^{c \times (d_b+1)}]$ is defined to extract body pose labels.
	
	%To simulate the missing data and the task of recovering the missing data, we set random projection mask $\Omega_{y}$ s.t observed entries $y_{ij} \in Y_{obs}, \forall (i,j) \in \Omega_{y}$. The unobserved entries are zeroed out. Since $\Omega_{y}$ is random, observed and unobserved entries are irregularly interleaved in $Y$. The objective is to predict the unobserved pose labels $Y_{\text{unobserved}}$ as part of the final predicted matrix $J^{*}$ in a classification task, utilizing all the visual features and the observed pose labels. 
	
	Matrix completion is an iterative method that attempts to fill in missing entries in a matrix, which in our context correspond to unobserved pose labels. For the purpose of the iterative scheme, the  unobserved pose labels can either be initialized by side information provided by external sources, or simply set to zeros. In this study, we take the first option by initializing the unobserved samples by sensor data.  The initial matrices for head and body poses are denoted by $\bm{J}_{0,h}$ and $\bm{J}_{0,b}$ respectively. The label matrix in $\bm{J}_{0,h}$, denoted by $\bm{Y}_h$, is further  divided into a training set $\bm{Y}_{\text{train},h}$ and a test set 
	%	\HH{I don't know why, but the convention is 'training set' and 'test set'. I have tried to edit where I see it but I might have missed some..} 
	$\bm{Y}_{\text{test},h}$. Similarly, the label matrix in $\bm{J}_{0,b}$ , denoted by $\bm{Y}_b$, is divided into $\bm{Y}_{\text{train},b}$ and $\bm{Y}_{\text{test},b}$. Each training set consists of observed labels, while the test set consists of unobserved labels. The training set and test set samples are interleaved, as shown in Figure \ref{mc}. In this study, training set labels are sampled from manual annotations and test set labels are initialized by sensor data, in the hope of achieving faster convergence. For the sake of brevity, the subsequent discussion will be explained for the  head pose matrix. The body pose matrix and its corresponding optimization formulation are analogous to those of the head pose matrix. 
	
	The following discussion outlines the proposed matrix completion method based on the aforementioned setting. The proposed method consists of three components: i) nuclear norm minimization, ii) temporal smoothing, and iii) head-body coupling.
	
	\subsection{Nuclear norm minimization} 
	Following the linear classifier assumption from \eqref{hetero matrix def}, \citet{GoldbergEtAl2010} have shown that the matrix $\bm{J}$ should be low rank. More concretely, the objective is to recover the missing pose labels such that the rank of the heterogeneous matrix  $\bm{J}$ is minimized. Rank minimization is a non-convex problem \cite{GoldbergEtAl2010}. However, \citet{CandesEtAl2010} showed that $\text{rank}(\bm{J})$ can be relaxed to its convex envelope which is the nuclear norm, $\|	 {\bm{J}}\|	_*$, i.e.
	\begin{equation}
	\text{rank} (\bm{J})  \approx \|	 {\bm{J}}\|	_*.
	\end{equation}
	The optimization problem then becomes a minimization of the nuclear norm of $\bm{J}$. 
	% \begin{equation}
	% J^h_*= \arg \min_{J^h}  \|	 {J^h}\|	_*
	%\end{equation}
	
	\subsection{Temporal smoothing} 
	If samples in the heterogeneous matrix are temporally sorted, one can take advantage of the temporal structure between the columns. Pose labels are to a certain extent, temporally smooth, as poses are not expected to change drastically within a short time period. This can be seen as a column-wise regularization. Using the training set $\bm{Y}_{\text{train}}$, an interpolated time series of pose labels $\bm{\tilde{Y}} $ can be generated using an appropriate interpolation scheme to estimate the unobserved pose labels entirely based on temporal consideration. In the proposed method, Gaussian process regression (GPR) is chosen as the interpolation scheme. Also known as Kriging, GPR has the same objective as other regression methods, which is to predict a value of a function at some point using a combination of observed values at other points. Rather than curve fitting using a polynomial function for instance, GPR assumes an underlying random process, more specifically a Gaussian process distribution  \cite{BachocEtAl2017}, from which the observed values are sampled. A new posterior distribution is computed based on the assumed (Gaussian process) prior and Gaussian likelihood functions \cite{WilliamsEtAl1998}. The Gaussian process prior is characterized by a covariance function which measures the similarity between data points; and thus the choice of a suitable covariance function is an essential component in GPR. For the purpose of this study, the covariance function is chosen to be the popular Radial-Basis Function (RBF) kernel. More details of Gaussian processes and Kriging can be found in \cite{RasmussenEtAl2005}. 
	
	Following this procedure, we denote  $\bm{Y}_{\text{GP}} \in \mathbb{R}^{c \times T}$ as the label matrix where the missing values are imputed by the prediction of GPR. After acquiring the interpolated labels, a new matrix $\bm{J}_{\text{GP}}$ is defined as
	\begin{equation}
	\bm{J}_{\text{GP}} = \begin{bmatrix} \bm{Y}_{\text{GP}} \\ \bm{X}  \\ \bm{1} \end{bmatrix}.
	\end{equation}
	A squared loss term $\|\bm{P} (\bm{J} - \bm{J}_{\text{GP}}) \|_F^2$ is introduced into the nuclear norm minimization problem for regularization to ensure that the predicted labels do not deviate drastically from the labels obtained as a result of temporal interpolation. The projection matrix $\bm{P}$ ensures that the loss is only considered over the pose labels.
	
	Note that GPR is an example of a regression method that works well in this setting. Alternative regression methods such Laplacian smoothing \cite{AlamedaPinedaEtAl2015}, piece-wise linear interpolation and polynomial regression can also be applied. Our justification of this choice follows in the discussion section in Section 7.

	%\textbf{Denoising} Additionally, it needs to be ensured that the observed entries don't deviate too much from their original values. Changes in the observed entries, both in terms of features and labels, are penalized by the additional square loss term. Hence, (3) is updated to
	%\begin{equation}
	%\underset{J^{*}}{\mathrm{argmin}} \;  \nu \|	 {J_{GP}}\|	_* + \lambda \| J^{*} - J_{\Omega_{y}} \|_{F}^2
	%\end{equation}
	%where $J_{\Omega_{y}}$ stands for the heterogeneous matrix with all visual features but only observed labels. 
	
	\subsection{Head and body coupling} 
	So far the formulation details the manipulation of HPE and BPE matrices separately. In this section we jointly consider the two matrices as they are related. Previous research by \citet{ChenEtAl2011joint, AlamedaPinedaEtAl2015, VaradarajanEtAl2018} has shown that coupling HPE and BPE is advantageous for improving accuracy. The proposed formulation also captures the physical constraints between head and body poses. Since head and body pose are jointly estimated, this relation fits in nicely as an additional regularization to the optimization problem. It is reasonable to model that head and body poses cannot be too different at any given time step. Though hinge loss would probably be more appropriate, the relation can also be captured by squared loss, for the ease of analytical derivation and numerical optimization. The regularization term can therefore be written as $\|\bm{P}_h \bm{J}_h - \bm{P}_b \bm{J}_b \|_F^2$. 
	
	\subsection{Optimization problem} 
	To summarize, the entire optimization problem, considering all the regularizations and indicating terms associated with both head and body (described in Section 4.1-4.3) is given by
	\begin{equation}
	\begin{aligned}\label{completeFormula}
	\bm{J}_{h *}, \bm{J}_{b*} =& \arg \min_{\bm{J}_{h},\bm{J}_{b}}  \nu_{h}  \|	 {\bm{J}_h}\|	_*+ \nu_{b}  \|	 {\bm{J}_b}\|	_* \\
	&+ \frac{\lambda_{h}}{2} \| \bm{P}_h (\bm{J}_{h} - \bm{J}_{\text{GP},h}))\|_{F}^2 + \frac{\lambda_{b}}{2} \| \bm{P}_b (\bm{J}_{b} - \bm{J}_{\text{GP},b})\|_{F}^2 \\
	&+ \frac{\mu}{2}  \|\bm{P}_h \bm{J}_h - \bm{P}_b \bm{J}_b \|_F^2 ,
	\end{aligned}
	\end{equation}
	where $\nu_{h}$, $\nu_{b}$, $\lambda_{h}$, $\lambda_{b}$, and  $\mu$ are weights that control the trade-off between the different terms. 
	The equation in \eqref{completeFormula} can be solved iteratively by an adapted Alternating Direction Method of Multipliers (ADMM) proposed by \citet{BoydEtAl2011} and \citet{AlamedaPinedaEtAl2015} to jointly solve the minimization problem for the head and body pose matrices. We adopt the aforementioned algorithm that starts with the construction of the augmented Lagrangian, similar to the classical ADMM \cite{EcksteinEtAl2012}. The augmented Lagrangian of the optimization problem in \eqref{completeFormula} is given by
	\begin{equation}
	\begin{aligned}\label{completeLagrangian}
	\mathcal{L} = \nu_{h} &  \|	 {\bm{J}_h}\|	_*+ \nu_{b}  \|	 {\bm{J}_b}\|	_* \\
	&+ \frac{\lambda_{h}}{2} \| \bm{P}_h (\bm{K}_{h} - \bm{J}_{\text{GP},h} )\|_{F}^2 + \frac{\lambda_{b}}{2} \| \bm{P}_b (\bm{K}_{b} - \bm{J}_{\text{GP},b})\|_{F}^2 \\
	&+ \frac{\mu}{2}  \|\bm{P}_h \bm{K}_h - \bm{P}_b \bm{K}_b \|_F^2  \\
	& + \frac{\phi_{h}}{2}  \| \bm{K}_h - \bm{J}_h \|_F^2  +  \frac{\phi_{b}}{2}  \| \bm{K}_b - \bm{J}_b \|_F^2 \\
	& + \langle \bm{M}_{h} , \bm{J}_{h}- \bm{K}_{h} \rangle + \langle \bm{M}_{b} , \bm{J}_{b}- \bm{K}_{b} \rangle,
	\end{aligned}
	\end{equation}
	where $\bm{K}_{h}$ and $\bm{K}_{b}$ are auxiliary variables that allow the decoupling of the optimization of $\bm{J}_h$ and $\bm{J}_b$; and $\bm{M}_h$ and $\bm{M}_b$ are Lagrange Multiplier matrices. The inner product of the two terms is denoted by $\langle  \: \cdot \: , \: \cdot  \: \rangle$ . The update rules are similar to those of the ADMM with scaled dual variables \cite{BoydEtAl2011}. In this context, the update rules at the $k$-th iteration are given by
	\begin{equation}
	\begin{aligned} \label{Jequations}
	(\bm{J}_{h}^{k+1}, \bm{J}_{b}^{k+1}) = \arg &\min_{\bm{J}_h^{k}, \bm{J}_b^{k}} \nu_{h}   \|	 {\bm{J}_h^{k}}\|	_*+ \nu_{b}  \|	 {\bm{J}_b^{k}}\|	_* \\
	% 	&+\frac{ \mu}{2}  \|\bm{P}_h \bm{K}_h^{k} - \bm{P}_b \bm{K}_b^{k} \|_F^2  \\
	& + \frac{\phi_{h}}{2}  \| \bm{K}_{h}^{k} - \bm{J}_h^{k} \|_F^2  + \frac{ \phi_{b}}{2}  \| \bm{K}_{b}^{k} - \bm{J}_b^{k} \|_F^2 \\
	& + \langle \bm{M}_{h}^{k} , \bm{J}_{h}^{k}- \bm{K}_h^{k} \rangle + \langle \bm{M}_{b}^{k} , \bm{J}_{b}^{k}- \bm{K}_{b}^{k} \rangle
	\end{aligned}
	\end{equation}
	
	\begin{equation}
	\begin{aligned} \label{Kequations}
	(\bm{K}_{h}^{k+1}, \bm{K}_{b}^{k+1}) = \arg &\min_{\bm{K}_h^{k}, \bm{K}_b^{k}} \frac{ \lambda_{h}}{2} \| \bm{P}_h (\bm{K}_{h}^{k} - \bm{J}_{\text{GP},h}) \|_{F}^2 \\
	&+\frac{ \lambda_{b}}{2} \| \bm{P}_b (\bm{K}_{b}^{k} - \bm{J}_{\text{GP},b} )\|_{F}^2 \\
	&+ \frac{\mu}{2}  \|\bm{P}_h \bm{K}_h^{k} - \bm{P}_b \bm{K}_b^{k} \|_F^2  \\
	& + \frac{\phi_{h}}{2}  \| \bm{K}_{h}^{k} - \bm{J}_{h}^{k+1} \|_F^2  \\
	&+  \frac{\phi_{b}}{2}  \| \bm{K}_{b}^{k} - \bm{J}_{b}^{k+1} \|_F^2 \\
	& + \langle \bm{M}_{h}^{k} , \bm{J}_{h}^{k+1}- \bm{K}_{h}^{k} \rangle + \langle \bm{M}_{b}^{k} , \bm{J}_{b}^{k+1}- \bm{K}_{b}^{k} \rangle
	\end{aligned}
	\end{equation}
	
	\begin{equation}
	\bm{M}_{h}^{k+1} = \bm{M}_{h}^{k} + \phi_{h} (\bm{J}_{h}^{k+1}- \bm{K}_{h}^{k+1})
	\end{equation}
	\begin{equation}
	\bm{M}_{b}^{k+1} = \bm{M}_{b}^{k} + \phi_{b} (\bm{J}_{b}^{k+1}- \bm{K}_{b}^{k+1})
	\end{equation}
	More derivation and implementation details can be found in Appendix \ref{appendix}. 
	
	%It is worth noting the potential intricacies involving the stopping criterion. In this formulation, we ensure absolute convergence in both head and body optimization, rather than prioritizing body over head, as previously done. 
	
	\section{Experimental Setup}
	This section provides a brief introduction of the SALSA dataset that was used to obtain the experimental results, and an overview of the experimental conditions. 
	\subsection{SALSA Dataset}
	The SALSA dataset is captured at a social event that consists of a poster presentation session and a mingling event afterwards, involving 18 participants. It is a multimodal dataset that includes video recordings from a multi-view surveillance camera (4 cameras) network, binary proximity sensor data acquired from sociometric badges worn by the participants, and audio recordings of each participant acquired by a microphone embedded in the sociometric badges. For this study, we only focus on the video recordings of the poster presentation session. Ground truth labels of head and body pose of each participant were manually annotated every 3 seconds. There are in total 645 ground truth annotations for each head and body pose of each participant. The authors of \cite{AlamedaPinedaEtAl2015} also inferred head pose from microphone data and body pose from infrared proximity sensor data, independent from the video modality. These are considered as "soft" labels and further details of their extraction can be found in \cite{AlamedaPinedaEtAl2015} and are provided as part of the dataset.
	
	\subsection{Experimental Conditions}
	We used the provided Histogram of Gradients (HOG) visual features for head and body crops of each participant from the SALSA dataset poster session \cite{AlamedaPinedaEtAl2015}. Similar to \citet{AlamedaPinedaEtAl2015}'s approach, visual features from the four cameras are concatenated and PCA was performed to keep 90\% of the variance. This results in a 100 dimensional feature vector. Training data are the observed labels and test data are the unobserved labels to be predicted. In a transductive learning setting, it is conventional to have both the training data and test data available during training. Because the objective is to predict labels for the unobserved entries only and not generalize to further unseen data, weights are not explicitly learned. Training data and test data partitions are defined by a random projection mask to simulate random sampling over labels. Because of this randomness, training and test data are interleaved and we take advantage of this inherent structure in our formulation. Note that because of the same reason, all our experiments are conducted 10 times and results are averaged to mediate the random projection mask causing overestimation or underestimation of prediction accuracies. Additionally, the sample diversity (i.e. class distribution) is different among participants. Hence, a randomly created projection mask is rejected if it results in low sample diversity in the training set. The hyper-parameters in (6) are optimized using Bayesian optimization with 5-fold cross validation.
	
	\subsection{Implementation Details}
	Similar to the authors of \cite{AlamedaPinedaEtAl2015}, we assume visual features from each participant are available at any given time step. Unlike in previous approach \cite{AlamedaPinedaEtAl2015} where the inferred "soft" labels are used as part of the training set, our experiments only use samples that were manually annotated to construct the training set. It is unclear if the experiments reported by \citet{AlamedaPinedaEtAl2015} used additional unlabeled samples along with the manual annotations and "soft" labels in their model during training. 
	
	Since we were not able to clarify ambiguities in the description of the experimental setup in the former formulation \cite{AlamedaPinedaEtAl2015}, we made the following decisions regarding the experimental setting.
	% the aforementioned experimental conditions in this study are our best interpretation of what was done previously and any possible variations from \cite{AlamedaPinedaEtAl2015} are to make this study more self-contained. 
	In this study, we construct the training and test sets from only samples that are manually annotated in the SALSA dataset. Since the quality of the "soft" labels was not quantitatively assessed by \citet{AlamedaPinedaEtAl2015}, in our case, it also makes sense for us to avoid training using "soft'' labels so we can more clearly see the effect of our proposed approach independently of the influence of training with weak labels. In our experiments, although "soft" labels are not considered as part of the observed samples, they are only used as initializations of unobserved samples in order to reach faster convergence.
	%	\HH{Didn't you also get a 2 percent increase in performance on average?}
	Note that columns of the matrix which are initially populated with soft labels are subject to immediate changes after being fed as inputs to the optimization problem.
	
	\section{Results}
\begin{figure}
		\includegraphics[width=\columnwidth]{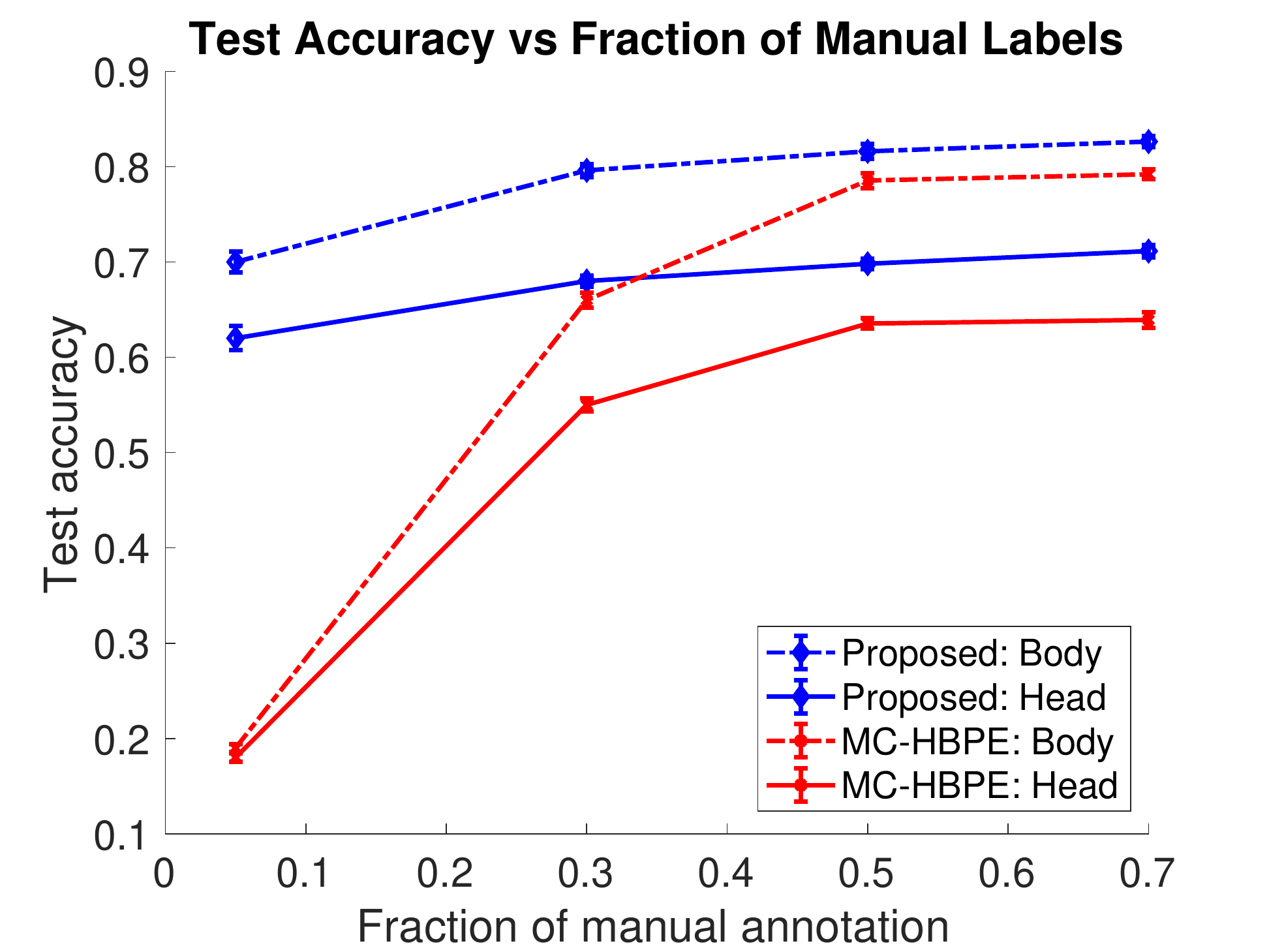}
		\caption{Test accuracy of HPE and BPE using MC-HBPE \cite{AlamedaPinedaEtAl2015} and the proposed method. Error bars indicate the standard deviations of results for each fraction of manual annotation.}\label{fig}
	\end{figure}

	The heterogeneous matrix for head and body are initialized with the same fraction of ground truth labels as training data, though their respective random projection masks are different. Figure \ref{fig} shows the test accuracy, which is the prediction accuracy over unobserved labels, against different fractions of manual annotation used for training. The proposed method is compared against the state-of-the-art matrix completion based HBPE method by  \citet{AlamedaPinedaEtAl2015}. As shown in Figure \ref{fig}, the proposed method is drastically superior compared to the state-of-the-art matrix completion by \citet{AlamedaPinedaEtAl2015}, especially at very low fraction of manual annotations.

	The difference in performance of both methods is accredited to a simple numerical phenomenon. One of the major differences between the proposed method and the method by \citet{AlamedaPinedaEtAl2015} is the temporal smoothing scheme. In the latter, the authors employed Laplacian smoothing to ensure temporal consistency over the pose estimates. While it is a reasonable choice for smoothing based on local information, GPR in contrast provides smoothing by exploiting a more global context based on only a few data points. By fitting sparse data points in the functional space, GPR is known to better recover nonlinear patterns and longer timescale trends compared to polynomial interpolation, and especially Laplacian smoothing. As a result, it provides a good accuracy even when only 5 \% of the manual labels are available as training data. Additionally, person-wise HBPE results for all 18 participants at 5 \% manual annotation using the two methods is reported in Table ~\ref{tablePersonWise}.
	
	% \usepackage{booktabs}
	% \usepackage{multirow}
	\begin{table*}[t]
		\begin{center}
			% \usepackage{booktabs}
			\begin{tabular}{@{}lllllll@{}}
				\toprule
				\begin{tabular}[c]{@{}l@{}}Manual \\ Annotation: 5 \%\end{tabular} & \multicolumn{2}{c}{MC-HBPE{[}2{]}} & \multicolumn{2}{c}{Proposed}    & \multicolumn{2}{c}{\begin{tabular}[c]{@{}c@{}}Labels diversity \\ (Entropy)\end{tabular}} \\ \midrule
				& HPE mean (std)  & BPE mean (std)   & HPE mean (std) & BPE mean (std) & Head                                        & Body                                        \\
				Person 1 {[}119{]}                                                 & 0 (0)           & 0 (0)            & 0.49 (2.9e-2)  & 0.57 (5.1e-2)  & 1.19                                        & 1.14                                        \\
				Person 2 {[}132{]}                                                 & 0.06 (2.3e-3)   & 0 (0)            & 0.39 (1.0e-2)  & 0.84 (1.8e-2)  & 1.32                                        & 0.48                                        \\
				Person 3 {[}140{]}                                                 & 0.63 (3.0e-2)   & 0.67 (4.1e-2)    & 0.77 (1.9e-2)  & 0.82 (3.2e-2)  & 1.51                                        & 1.29                                        \\
				Person 4 {[}169{]}                                                 & 0.02 (1.6e-3)   & 0.01 (3.5e-3)    & 0.85 (3.6e-2)  & 0.86 (2.7e-2)  & 1.20                                        & 1.10                                        \\
				Person 5 {[}177{]}                                                 & 0.13 (2.9e-3)   & 0.13 (1.2e-2)    & 0.53 (5.4e-2)  & 0.60 (6.1e-2)  & 1.84                                        & 1.79                                        \\
				Person 6 {[}180{]}                                                 & 0.44 (1.6e-2)   & 0.39 (1.7e-2)    & 0.65 (4.0e-2)  & 0.75 (4.6e-2)  & 1.72                                        & 1.56                                        \\
				Person 7 {[}216{]}                                                 & 0.17 (6.6e-2)   & 0.17 (3.1e-2)    & 0.56 (3.3e-2)  & 0.48 (7.5e-2)  & 1.77                                        & 1.90                                        \\
				Person 8 {[}238{]}                                                 & 0.01 (5.2e-4)   & 1.5e-4 (5.2e-4)  & 0.82 (1.1e-2)  & 0.88 (2.3e-2)  & 0.60                                        & 0.37                                        \\
				Person 9 {[}241{]}                                                 & 0.34 (4.1e-3)   & 0.57 (4.9e-3)    & 0.63 (7.7e-2)  & 0.70 (6.9e-2)  & 1.57                                        & 1.59                                        \\
				Person 10 {[}261{]}                                                & 0.09 (2.6e-3)   & 0.12 (2.8e-3)    & 0.69 (1.6e-2)  & 0.85 (2.9e-2)  & 1.39                                        & 1.21                                        \\
				Person 11 {[}262{]}                                                & 0.13 (1.4e-3)   & 0.01 (1.7e-3)    & 0.60 (4.7e-2)  & 0.69 (5.6e-2)  & 1.56                                        & 1.50                                        \\
				Person 12 {[}267{]}                                                & 0.13 (6.8e-3)   & 0.03 (8.4e-3)    & 0.81 (1.9e-2)  & 0.82 (1.8e-2)  & 1.01                                        & 0.96                                        \\
				Person 13 {[}286{]}                                                & 0 (0)           & 0 (0)            & 0.68 (2.4e-2)  & 0.75 (3.7e-e)  & 1.66                                        & 1.60                                        \\
				Person 14 {[}307{]}                                                & 0.09 (2.7e-2)   & 0.12 (3.7e-2)    & 0.37 (4.4e-2)  & 0.46 (7.7e-2)  & 1.88                                        & 1.79                                        \\
				Person 15 {[}313{]}                                                & 0 (0)           & 0 (0)            & 0.57 (6.0e-2)  & 0.65 (4.7e-2)  & 1.16                                        & 1.06                                        \\
				Person 16 {[}350{]}                                                & 0.03 (2.7e-3)   & 0.03 (2.9e-2)    & 0.69 (7.6e-2)  & 0.69 (7.0e-2)  & 1.23                                        & 1.23                                        \\
				Person 17 {[}351{]}                                                & 0.13 (4.9e-2)   & 0.25 (4.1e-2)    & 0.52 (3.7e-2)  & 0.51 (4.3e-2)  & 1.74                                        & 1.74                                        \\
				Person 18 {[}353{]}                                                & 0.13 (2.2e-2)   & 0.20 (8.2e-3)    & 0.55 (6.1e-2)  & 0.72 (7.2e-2)  & 1.41                                        & 1.12                                        \\ 
				\bottomrule
			\end{tabular}
		\end{center}
		\begin{tablenotes}
			\item $^*$ $[\cdot]$ indicates the person ID encoding provided in the SALSA dataset.
		\end{tablenotes}
		\caption{Person-Wise HBPE results using MC-HBPE \cite{AlamedaPinedaEtAl2015} and the proposed method.  Difficulty of HBPE for each person is captured quantitatively in labels diversity measured by labels entropy. }
		\label{tablePersonWise}
	\end{table*}
	
	During social events and in free-standing conversation groups, we expect head pose to change more than body poses and that these changes are fine-grained. Hence, it is reasonable to conclude that head poses are harder to predict compared to body poses; and it is reflected in the observation that test accuracies for head pose estimates are lower than test accuracies for body pose estimates from both the methods. This can be further analyzed by computing information entropy to illustrate the distribution of the ground truth labels used in this study. The equation for calculating entropy is  given by
	\begin{equation}
	H= - \sum_{i=1}^{c} P_i \text{log} P_i,
	\end{equation}
	where $H$ is the information entropy measure of a set of samples and $P_i$ is the proportion of ground truth labels in the $i^{\text{th}}$ class. For unbiased 8 class label distribution (i.e. uniform distribution), the maximum entropy value is approximately 2.08. The entropy of head pose labels averaged over all participants is 1.43 with standard deviation 0.33. The entropy of body pose labels averaged over all participants is 1.3 with standard deviation 0.43. Therefore, head pose diversity is slightly higher than that of the body pose, which partially justifies the reasoning that head pose labels are more difficult to accurately predict than body pose labels. However, the GPR-based proposed method still manages to achieve significantly higher test accuracies for head pose estimates compared to the method by \citet{AlamedaPinedaEtAl2015}.  
	%	 Since the results obtained from MC-HBPE by \citet{AlamedaPinedaEtAl2015} are less conclusive compared to those of the proposed method, we perform further analysis by relating label distribution to performance on the latter set of results. We find that HBPE performance are negatively correlated with labels diversity (-0.53 for HPE and -0.71 for BPE), which means that the lower the diversity, the better the performance. This is an explainable phenomenon given the context of the model. 
	
	It is worth noting that in this study, we sample training data from manual labels, whereas in the experimental setup by \citet{AlamedaPinedaEtAl2015}, "soft" labels acquired from wearable devices are also used as part of training data. Experiments were also conducted where the "soft" labels provided in the dataset are included as part of the training data. However, no desirable results can be obtained. As a reference, using the same approach as that of \cite{AlamedaPinedaEtAl2015} at 50\% training data partition with 5\% manual annotations and 95 \% "soft" labels, we obtained 14\% and 16 \% for HPE and BPE respectively, compared to the reported 57\% and 60\% \cite{AlamedaPinedaEtAl2015}.

	\section{Discussion}
	
	In our proposed method, GPR performs fitting over the head and body pose estimates separately, which loosens the head and body coupling constraint to a certain extent. Though there is still point to point coupling between head pose and body pose at each time step, the head poses and body poses are each separately governed by their own trend which should be less sensitive to noise from the other. Coupling that is too tight may artificially enforce head and body pose to be the same which may not reflect the reality when it comes to small group interactions. This implicit benefit from recovering nonlinearities independently should provide rich information to study human behavior in groups. 
	%\HH{I think you need to link this back to the results. e.g. the the improvement in head pose estimation is more than the improvement in body pose estimation even when the fraction of manual annotation increases to a point where the effect of GPR is no longer better than the Laplacian smoothing approach.}
	
	%	\HH{I am stopping reading here as it seems that this part hasn't been changed since my previous comments.}
	Since wearable sensors are known to provide noisy information, not all "soft" labels can be seen to have the same quality as ground truth labels. It would be ideal to add high quality "soft" labels to training data and if they are as high quality as manual labels, they can further benefit and improve HBPE in a multimodal setting, as opposed to a single video modality. However, this prior knowledge would need to be obtained beforehand. Because the proposed formulation gives robust performance with small number of manual annotations without the use of any "soft" labels, it provides a good baseline and ground for comparison for further investigation of the quality of labels derived from wearable sensors. 
	
	While the highlight of this formulation is to predict the classification of unobserved labels based on a very small number of observed labels, the model does not extend to predicting further unseen data since the weights are not explicitly recovered. When an observed label becomes available to be included, the full model needs to be run again. One of the computational bottlenecks is Gaussian process regression, which has $O(n^3)$ time complexity that makes it infeasible to scale up for large quantities of data. Another computational bottleneck is the singular value decomposition (SVD) in solving the optimization problem using ADMM (see Appendix \ref{appendix}).
	
	\section{Conclusion}
	This work focuses on estimating head and body poses in crowded social scene scenario using Gaussian process regression and head and body coupling as a regularization term in a matrix completion setting. The model's premise is to predict head and body pose labels as an 8-class classification problem in a transductive learning setting. The model is able to predict a relatively large percentage of pose labels in large continuous time segment (average 20 samples gap length, approximately 1 minute in real time) and implicitly recover the nonlinearity within the data using only a small fraction of samples as training data. The proposed model has shown to be effective on the challenging SALSA dataset and achieved desirable results of ~62 \% accuracy on head pose estimation and ~70\% accuracy on body pose estimation using only ~5\% of the samples as training data, showing superior performance over the state-of-the-art. 
	
	Future work on improving HBPE includes integrating wearable sensor data as regularization terms towards a truly multimodal approach. Rather than using appearance based HOG features, visual features could also be extracted using a CNN pre-trained on large image databases and fine-tuned on the SALSA dataset. Additionally, it would be interesting to assess the performance of the proposed method on different, but equally challenging datasets, such as the MatchNMingle dataset \cite{Cabrera-QuirosEtAl2018}. Further analysis of HBPE performance with respect to participants' role in the social scenarios in question and their pose diversity may lend deeper insights to fine-grained head and body movements in group interactions. 
	%\begin{itemize}
	%	\item recap on how this method can handle more missing data and that it got desirable results on a subset of the SALSA dataset
	%	\item mention future work in making this a generalized multimodal framework
	%	\item approx. 0.3 column
	%\end{itemize}
	
	\section{Acknowledgment}
	The authors thank Xavier Alameda-Pineda for sharing data and implementation of his previous research \cite{AlamedaPinedaEtAl2015}. 
	%This research was partially funded by the Netherlands Organization for Scientific Research (NWO) under project number 639.022.606.

	\bibliographystyle{ACM-Reference-Format}
	\bibliography{MCbib}
	\appendix
	
	\section{Derivations of ADMM}
	\label{appendix}
	%The objective of solving \eqref{completeLagrangian} is to obtain $\bm{J}^{k+1,h}$, $\bm{J}^{k+1,b}$, $\bm{K}^{k+1,h}$, and $\bm{K}^{k+1,b}$. 
	To separate head and body expressions, at $k^{\text{th}}$ iteration, the optimization problem \eqref{Jequations} can be split into
	\begin{equation}
	\begin{aligned} \label{Jhead}
	\bm{J}_h^{k+1}  = \arg \min_{\bm{J}_h^{k}}  \nu_h  \|	 {\bm{J}_h^{k} }\|	_*  + \frac{\phi_{h}}{2}  \| \bm{K}_{h}^{k} - \bm{J}_h^{k} \|_F^2  
	+ \langle \bm{M}_{h}^{k} , \bm{J}_{h}^{k} - \bm{K}_{h}^{k} \rangle
	\end{aligned} 
	\end{equation} and
	\begin{equation}
	\begin{aligned} \label{Jbody}
	\bm{J}_b^{k+1}  = \arg \min_{\bm{J}_b^{k}}  \nu_b \|	 {\bm{J}_b^{k}}\|	_*  + \frac{\phi_{b}}{2}  \| \bm{K}_{b}^{k} - \bm{J}_b^{k} \|_F^2  
	+ \langle \bm{M}_{b}^{k} , \bm{J}_{b}^{k} - \bm{K}_{b}^{k} \rangle .
	\end{aligned} 
	\end{equation} 
	Simplifying and manipulating \eqref{Jhead}, we obtain 
	\begin{equation}
	\begin{aligned} \label{simp}
	\bm{J}_h^{k+1}  = &\arg \min_{\bm{J}_h^{k}}  \nu_h \|{\bm{J}_h^{k}}\|	_* \\
	&+ \frac{\phi_{h}}{2}  [\langle \bm{K}_{h}^{k},  \bm{K}_{h}^{k} \rangle - 2 \langle \bm{K}_{h}^{k} , \bm{J}_{h}^{k} \rangle  + \langle \bm{J}_{h}^{k} \bm{J}_{h}^{k} \rangle ]\\
	&+ \langle \bm{M}_{h}^{k} , \bm{J}_{h}^{k} \rangle- \langle \bm{M}_{h}^{k} , \bm{K}_{h}^{k} \rangle  \\
	&+ \frac{1}{2\phi_h}  \langle \bm{M}_{h}^{k} , \bm{M}_{h}^{k} \rangle -   \frac{1}{2\phi_h}  \langle \bm{M}_{h}^{k} , \bm{M}_{h}^{k} \rangle.\\ 
	\end{aligned} 
	\end{equation} 
	Equation \eqref{simp} can be arranged as
	\begin{equation}
	\begin{aligned} \label{equiJhead}
	\bm{J}_h^{k+1} =& \arg\min_{J_h} \frac{\nu_{h}}{\phi_{h}} \|\bm{J}_{h}\|_{*} + 
	\frac{1}{2}\left\| \frac{1}{\phi_{h}} \bm{M}_h^{k} + \bm{J}_{h}-\bm{K}_{h}^{k} \right\|_F^2 -   \frac{1}{2\phi_h}  \langle \bm{M}_{h}^{k} , \bm{M}_{h}^{k} \rangle.
	\end{aligned}
	\end{equation}  
	%Following \cite{CaiEtAl2010}, the solutions of Equation \eqref{equiJhead}  is obtained using the singular value thresholding algorithm. 
	The last term in Equation \eqref{equiJhead} results in a scalar constant and does not affect the nature of optimization. The solution to minimization problem \eqref{equiJhead} was derived by \citet{CaiEtAl2010} and \citet{AlamedaPinedaEtAl2015}, and is given by
	\begin{equation}
	\begin{aligned} \label{svdhead}
	\bm{J}_h^{k+1} = \bm{U}_h S_{\frac {\nu_h}{\phi_h}} (\bm{D}_h)\bm{V}_{h} ^\top
	\end{aligned} ,
	\end{equation} 
	where the $\bm{U}_h$, $\bm{D}_h$, and  $\bm{V}_h$ are obtained from singular value decomposition (SVD) of matrix $\bm{K}_h^{k} - \frac{1}{\phi_h} \bm{M}_h^{k}$
	\begin{equation}
	[\bm{U}_h, \bm{D}_h, \bm{V}_h]  = \text{SVD} \left(\bm{K}_h^{k} - \frac{1}{\phi_h} \bm{M}_h^{k}\right)
	\end{equation}
	and where the shrinkage operator is given by
	\begin{equation}
	\label{shrinkage}
	%		S_{\frac {\nu_h}{\phi_h}} (x) = \text{max} \left(x-\frac {\nu_h}{\phi_h},0\right) 
	S_{\lambda} (x) = \text{max} \left(x-\lambda,0\right) 
	\end{equation}
	and is applied element-wise to the diagonal matrix of singular values $\bm{D}_h$.  The derivations can be similarly extended for body pose estimation matrix and the solution is given by
	\begin{equation}
	\begin{aligned} \label{svdbody}
	\bm{J}_b^{k+1} = \bm{U}_b S_{\frac {\nu_b}{\phi_b}} (\bm{D}_b)\bm{V}_{b} ^ \top
	\end{aligned} .
	\end{equation} 
	
	For the second step in the optimization problem \eqref{Kequations}, we define the row-vectorization form of the matrices $\bm{K}_h$ and $\bm{K}_b$ as $\bm{k}_h = \text{vec}(\bm{K}_h)$ and $\bm{k}_b = \text{vec}(\bm{K}_b)$ respectively. The row vectorization notation extends to other matrices in \eqref{Kequations} similarly. Derivatives of the objective function in \eqref{Kequations} with respect to $\bm{k}_{h}$ and $\bm{k}_{b}$  are given by
	\begin{equation}
	\label{derivative_kh}
	\pdv{\mathcal{L}}{\bm{k}_h} = \lambda_{h}(\bm{k}_h-\bm{j}_{GP,h})+\mu \bm{P}_{h}^ \top (\bm{P}_h \bm{k}_h- \bm{P}_b \bm{k}_b)+\phi_h (\bm{k}_h-\bm{j}_{h}^{k+1})-\bm{m}_{h}^{k},
	\end{equation}
	and
	\begin{equation}
	\label{derivative_kb}
	\pdv{\mathcal{L}}{\bm{k}_b} = \lambda_{b}(\bm{k}_b-\bm{j}_{GP,b})+\mu \bm{P}_{b}^ \top (\bm{P}_b \bm{k}_b- \bm{P}_h \bm{k}_h)+\phi_b (\bm{k}_b-\bm{j}_{b}^{k+1})-\bm{m}_{b}^{k} . 
	\end{equation}
	Equating this derivative to 0 results in a system of linear equations for $\bm{k}_{h}^{k+1}$ and $\bm{k}_{b}^{k+1}$ given by
	\begin{equation}
	\label{solveforkh}
	(\lambda_h + \mu \bm{P}_{h} ^\top \bm{P}_h + \phi_h) \bm{k}_{h}^{k+1} = \lambda_h \bm{j}_{GP,h} + \mu \bm{P}_{h}^ \top \bm{P}_b \bm{k}_b + \phi_h \bm{j}_{h}^{k+1}+ \bm{m}_{h}^{k}
	\end{equation}
	and
	\begin{equation}
	\label{solveforkb}
	(\lambda_b + \mu \bm{P}_{b}^ \top \bm{P}_b + \phi_b) \bm{k}_{b}^{k+1} = \lambda_b \bm{j}_{GP,b} + \mu \bm{P}_{b}^ \top \bm{P}_h \bm{k}_h + \phi_b \bm{j}_{b}^{k+1}+ \bm{m}_{b}^{k} .
	\end{equation}
	Hence, these two equations can be easily solved using standard solvers based on LU decomposition or iterative solvers such as conjugate gradient method to yield the minimizers $\bm{k}_{h}^{k+1}$ and $\bm{k}_{b}^{k+1}$. We can reshape the solved row vectors $\bm{k}_{h}^{k+1}$ and $\bm{k}_{b}^{k+1}$ back to matrix forms denoted by $\bm{K}_{h}^{k+1}$ and $\bm{K}_{b}^{k+1}$. Additionally, the system of linear equations \eqref{solveforkh} and \eqref{solveforkb} can be further simplified to give analytical solutions. For the sake of brevity, the reader is referred to the derivation by \citet{AlamedaPinedaEtAl2015} in their supplementary material.
\end{document}